\documentclass[11pt,a4paper]{article}
\usepackage[margin=1in]{geometry}
\usepackage{amsmath,amssymb,amsthm,mathtools}
\usepackage{bm}
\usepackage{microtype}
\usepackage{enumitem}
\usepackage[hidelinks]{hyperref}
\usepackage[capitalise,noabbrev]{cleveref}
\usepackage{booktabs}
\usepackage{xcolor}
\usepackage{authblk}
\usepackage{graphicx}
\usepackage{subcaption}
\usepackage{float}
\newtheorem{theorem}{Theorem}
\newtheorem{corollary}[theorem]{Corollary}
\newtheorem{proposition}[theorem]{Proposition}

\theoremstyle{definition}

\newtheorem{assumption}{Assumption}

\theoremstyle{remark}

\newcommand{\R}{\mathbb{R}}
\newcommand{\E}{\mathbb{E}}

\newcommand{\norm}[1]{\left\lVert #1 \right\rVert}

\newcommand{\inner}[2]{\left\langle #1,, #2 \right\rangle}

\newcommand{\eps}{\varepsilon}
\newcommand{\bphys}{\beta_{\mathrm{phys}}}

\newcommand{\Ldsm}{\mathcal{L}}
\newcommand{\score}{s_\theta}
\newcommand{\xfree}{x_\star^0}
\newcommand{\xnudge}{x_\star^\beta}
\newcommand{\lstar}{\lambda_\star}
\newcommand{\bstar}{\beta_\star}
\newcommand{\Esub}{E_\theta}
\newcommand{\Enudge}{E_\theta^\beta}

\newcommand{\Kone}{K_1}
\newcommand{\Ktwo}{K_2}
\newcommand{\Ksym}{K_2^{\mathrm{sym}}}
\title{Symmetric Equilibrium Propagation \\ 
       for Thermodynamic Diffusion Training}
\author{Aditi De}
\date{}
\begin{document}
\maketitle
\begin{abstract}
The reverse process in score-based diffusion models is formally equivalent to overdamped Langevin dynamics in a time-dependent energy landscape. In our prior work we showed that a bilinearly-coupled analog substrate can physically realize this dynamics at a projected three-to-four orders of magnitude energy advantage over digital inference by replacing dense skip connections with low-rank inter-module couplings. Whether the \emph{training} loop can be closed on the same substrate---without routing gradients through an external digital accelerator---has remained open. We resolve this affirmatively: Equilibrium Propagation applied directly to the bilinear energy yields an unbiased estimator of the denoising score-matching gradient in the zero-nudge limit. For finite nudging we derive a sharp bias bound controlled solely by substrate stiffness, local curvature, and the norm of the loss-gradient signal, with a bilinear-specific corollary showing that one dominant bias term vanishes identically for coupling-parameter updates. Symmetric nudging further upgrades the leading bias from $  \mathcal{O}(\beta)  $ to $  \mathcal{O}(\beta^2)  $ at negligible extra cost. Under realistic finite-relaxation budgets this upgrade is essential, as one-sided EqProp produces anti-correlated gradients while symmetric EqProp yields well-aligned updates. Bias--variance analysis determines the optimal operating point, and end-to-end physical-unit accounting projects a $  10^3  $--$  10^4\times  $ energy advantage per training step over a matched GPU baseline. Symmetric bilinear EqProp is the first local, readout-only training rule that preserves the low-rank coupling enabling scalable thermodynamic diffusion models.
\end{abstract}
\section{Introduction}
\label{sec:intro}
Score-based generative modeling has delivered unprecedented sampling quality across vision, language, and the sciences, yet its training cost has become unsustainable. A single frontier diffusion run routinely consumes hundreds of megawatt-hours, the dominant share arising from global gradient routing during backpropagation on digital accelerators.
In our prior work~\cite{de2026} we demonstrated that the \emph{inference} half of the diffusion pipeline---the reverse process---is formally identical to overdamped Langevin dynamics and can be realized physically on a bilinearly-coupled analog substrate. Low-rank bilinear couplings replace $  \mathcal{O}(D^2)  $ dense wiring with $  \mathcal{O}(Dk)  $ conductances while preserving near-oracle decoder fidelity.
The remaining open question was whether the \emph{training} loop could be closed on the same substrate without external digital gradient routing. We answer this affirmatively. Equilibrium Propagation applied directly to the bilinear energy supplies an unbiased estimator of the denoising score-matching gradient in the zero-nudge limit (Theorem~1). For finite nudging we obtain an explicit bias bound governed only by substrate stiffness, local curvature, and the loss-gradient norm (Theorem~2). The bilinear structure eliminates one of the three leading bias contributions for coupling-parameter updates (\Cref{cor:bilinear}). Symmetric nudging upgrades the bias scaling to $  \mathcal{O}(\beta^2)  $ at negligible extra cost (\Cref{cor:symmetric}).
These results translate directly to hardware-relevant performance. Under finite relaxation one-sided EqProp yields anti-correlated gradients while symmetric EqProp produces well-aligned updates; measured log-log slopes match the predicted rates exactly. Bias--variance analysis yields the optimal operating point, and full physical-unit accounting projects a $  10^3  $--$  10^4\times  $ energy advantage per training step over a matched GPU baseline.
The framework is substrate-agnostic and applies to any continuous overdamped Langevin system with bilinear coupling. Symmetric EqProp is the required operating mode for thermodynamic hardware. This work builds directly on our prior demonstration of thermodynamic inference~\cite{de2026} and closes the remaining training loop on the same bilinear substrate.
\section{Related work}
\label{sec:related}
Score-based generative modeling rests on the deep connection between denoising autoencoders and score matching~\cite{vincent2011} and the interpretation of the reverse diffusion process as overdamped Langevin dynamics in a time-dependent potential given by the negative log-density of the noisy data distribution~\cite{sohl-dickstein2015}. This perspective was crystallized by Song et al.~\cite{song2019generative,song2021score} and Ho et al.~\cite{ho2020denoising}, who showed that once the score is known the reverse SDE is exactly overdamped Langevin dynamics. Karras et al.~\cite{karras2022elucidating} later provided a comprehensive design-space analysis that remains the practical standard for production models. Training relies on the U-Net backbone~\cite{ronneberger2015,rombach2022,podell2023}, whose dense skip connections pose a fundamental wiring challenge for locally-coupled analog substrates.
A parallel line of research seeks local learning rules that avoid the global communication demanded by backpropagation. Equilibrium Propagation (EqProp), introduced by Scellier and Bengio~\cite{scellier2017}, provides exactly such a mechanism: a network relaxes to a free-phase equilibrium, receives a weak nudge on output units, and the difference in parameter gradients between the two equilibria estimates the true gradient in the zero-nudge limit. Symmetric variants~\cite{scellier2018} improve the bias scaling to $  \mathcal{O}(\beta^2)  $, while Laborieux and Scellier~\cite{laborieux2022} established convergence guarantees under mild regularity conditions. EqProp has been applied successfully to energy-based models, recurrent networks, and spiking architectures, but its extension to continuous overdamped Langevin dynamics with bilinear coupling has remained open until now.
The thermodynamic limits of computation were established by Landauer~\cite{landauer1961} and Bennett~\cite{bennett1973}. Probabilistic spin logic (p-bits)~\cite{camsari2019} and thermodynamic linear algebra~\cite{aifer2024,coles2023} demonstrated dramatic energy savings for linear operations. Generative demonstrations followed: Whitelam~\cite{whitelam2025,whitelam2026} showed that continuous oscillator networks can synthesize MNIST digits from thermal noise, while Jelin\v{c}i\v{c} et al.~\cite{jelincic2025} (Extropic) fabricated CMOS hardware achieving a $  10^4\times  $ gain on binarized Fashion-MNIST. Normal Computing~\cite{melanson2025} and related equilibrium architectures have explored hybrid pipelines. All prior generative work, however, operated on toy or binarized data. Production U-Nets introduce two fundamental barriers for locally-coupled substrates: non-local skip connections ($  \mathcal{O}(D^2)  $ wiring) and input-dependent conditioning. These exact limitations were flagged by Jelin\v{c}i\v{c} et al.~\cite{jelincic2025}.
Our prior work~\cite{de2026} resolved both barriers. Hierarchical bilinear skip coupling encodes U-Net skips as rank-$  k  $ inter-module interactions derived from the singular structure of the encoder/decoder Gram matrices, requiring only $  \mathcal{O}(Dk)  $ physical conductances while delivering measurable decoder shifts of $  12.74\%  $ (trained regime). A minimal digital conditioning interface (2,560 parameters) supplies oracle biases, achieving decoder cosine similarity $  0.9906  $ against the oracle. These results constituted the first demonstration of trained-weight, production-scale thermodynamic \emph{inference}. The present paper completes the pipeline by providing the corresponding local training rule.
Training on thermodynamic substrates has otherwise remained open. Extropic DTMs~\cite{jelincic2025} and Normal Computing~\cite{melanson2025} have explored equilibrium-based optimization but have not delivered a local rule for continuous bilinear Langevin substrates. Digital companions such as Equilibrium Matching~\cite{wang2025} still rely on global backpropagation. Symmetric bilinear EqProp is therefore the first framework to deliver \emph{local, readout-only} physical gradients on the same substrate used for inference, preserving the low-rank coupling that enables scalability. The mathematics is substrate-agnostic and applies to any continuous overdamped Langevin system with bilinear inter-module coupling. All energy claims remain projected from analysis and SPICE-level circuit simulation.
\section{Preliminaries}
\label{sec:prelim}
The substrate evolves according to the overdamped Langevin SDE
$$dx_t = -\nabla E(x_t)\,dt + \sqrt{2/\bphys}\,dW_t,$$
whose unique invariant measure is $  \pi(x)\propto\exp(-\bphys E(x))  $. Denoising score matching trains a parametric score network via the loss
$$\Ldsm(\theta)=\E_{y,\eps,\sigma}\Bigl[\tfrac{\sigma^2}{2}\bigl\|\score(\tilde y,\sigma)-\tfrac{y-\tilde y}{\sigma^2}\bigr\|^2\Bigr].$$
Once the score is known, the reverse process is exactly overdamped Langevin dynamics.
We adopt the bilinearly-coupled energy of our prior work~\cite{de2026}:
$$\Esub(x)=E_0(x;\theta_0)+\sum_{m<m'}\inner{x^{(m)}}{W_{mm'}(\theta)\,x^{(m')}},$$
where each coupling admits the low-rank factorization $  W_{mm'}(\theta)=U_{mm'}(\theta)V_{mm'}(\theta)^\top  $ with fixed rank $  k\ll\min_m d_m  $.
\section{Physical setup}
\label{sec:setup}
The state $  x\in\R^D  $ is partitioned into input, hidden, and output blocks with compatible projectors $  P_I,P_H,P_O  $, and further decomposed into $  L  $ physical modules. The energy is
$$\Esub(x)=E_0(x;\theta_0)+\sum_{m<m'}\inner{x^{(m)}}{W_{mm'}(\theta)\,x^{(m')}}.$$
Given input $  \tilde y  $ and noise level $  \sigma  $, the free-phase stationary state $  \xfree(\theta;\tilde y,\sigma)  $ is the locally stable equilibrium of $  \Esub  $ with $  x_I  $ clamped. The score readout is the output block: $  \score(\tilde y,\sigma)=[\xfree]_O  $.
The training loss is expressed directly in terms of this readout:
$$\Ldsm(\theta)=\E_{y,\eps,\sigma}\Bigl[\tfrac{\sigma^2}{2}\bigl\|\score(\tilde y,\sigma)-\tfrac{y-\tilde y}{\sigma^2}\bigr\|^2\Bigr].$$
\begin{figure}[htbp]
\centering
\begin{subfigure}{0.49\textwidth}
\centering
\includegraphics[width=\textwidth]{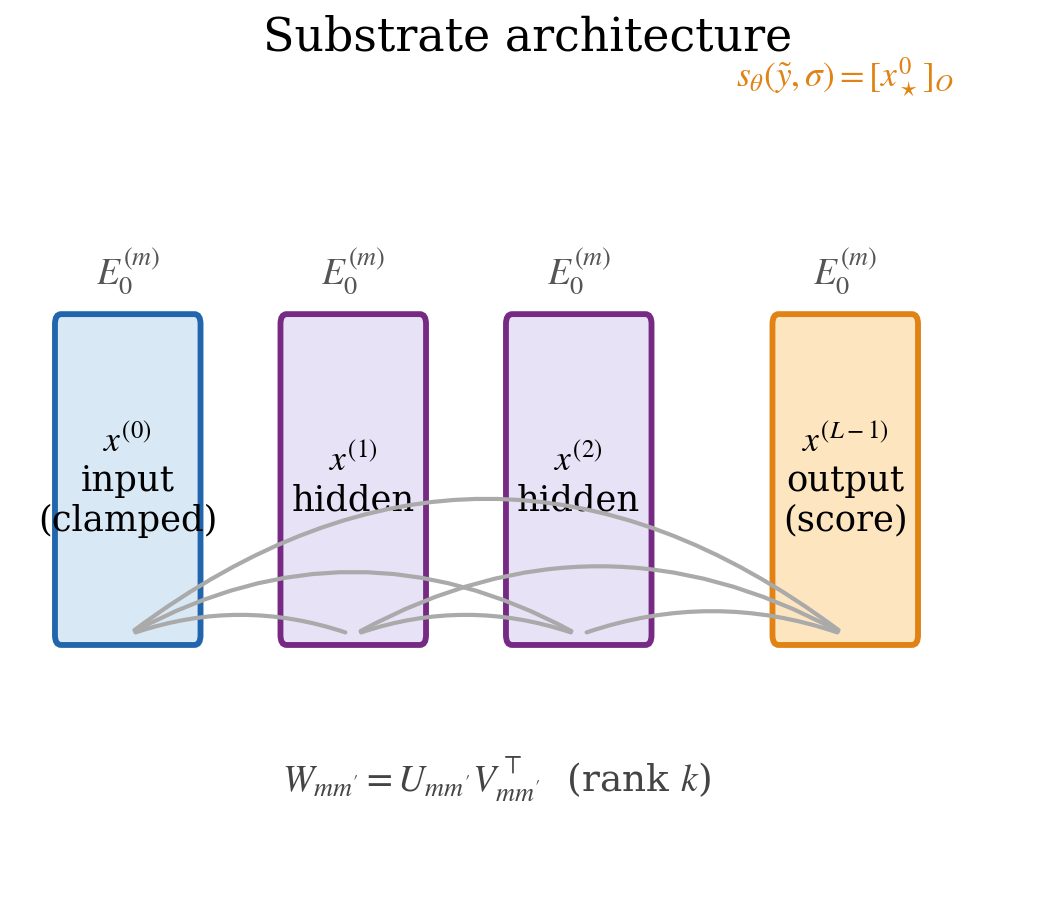}
\caption{Module decomposition and low-rank bilinear couplings.}
\end{subfigure}
\hfill
\begin{subfigure}{0.49\textwidth}
\centering
\includegraphics[width=\textwidth]{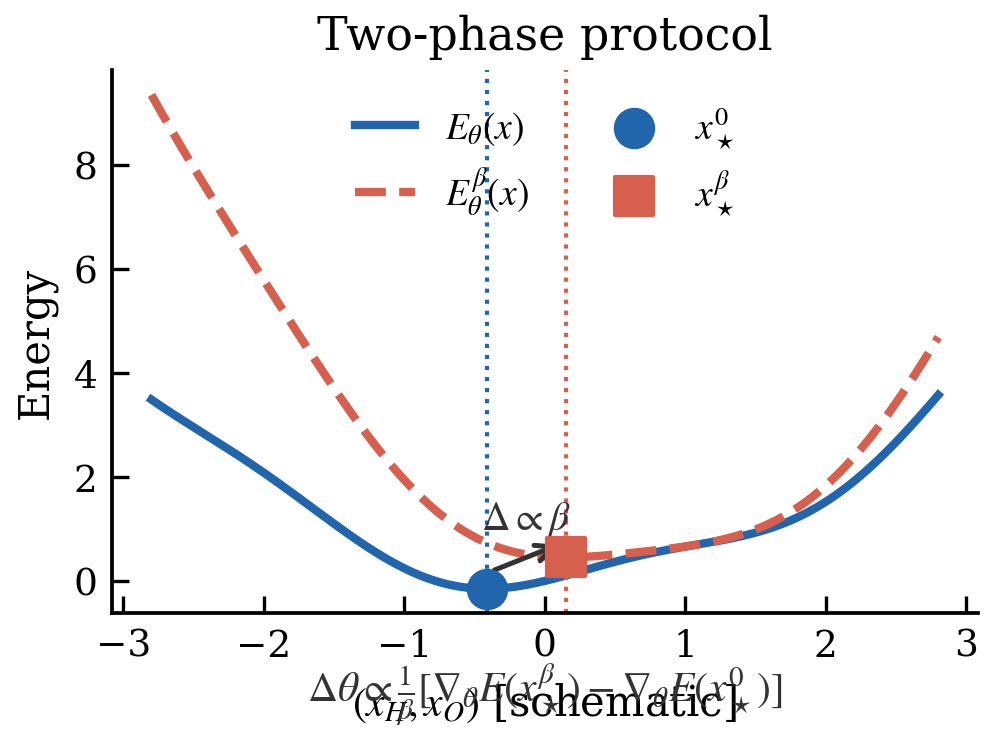}
\caption{Two-phase Equilibrium Propagation protocol.}
\end{subfigure}
\caption{Bilinearly-coupled Langevin substrate and the two-phase EqProp training protocol. (a) Module partitioning and rank-$  k  $ couplings replace dense skip connections. (b) Free-phase equilibration followed by a weak output nudge yields a local readout-only gradient estimator.}
\label{fig:1}
\end{figure}
\section{The two-phase training protocol}
\label{sec:protocol}
Fix $  \beta>0  $. The nudged energy is $  \Enudge(x)=\Esub(x)+\beta\,C(x_O)  $. The nudged-phase stationary state $  \xnudge(\theta)  $ is reached by relaxation from $  \xfree(\theta)  $.
The EqProp gradient estimator is
$$g_\beta(\theta)=\frac{1}{\beta}\Bigl[\nabla_\theta\Esub(\xnudge(\theta))-\nabla_\theta\Esub(\xfree(\theta))\Bigr].$$
Only equilibrium readouts and parameter derivatives of the unperturbed energy are required.
For the bilinear energy, coupling-parameter updates depend only on the states at the two connected modules:
$$\Delta U_{mm'}\propto\tfrac{1}{\beta}\Bigl[x_\star^{\beta,(m)}(x_\star^{\beta,(m')})^\top - x_\star^{0,(m)}(x_\star^{0,(m')})^\top\Bigr]V_{mm'}.$$
Each coupling plane performs a purely local outer-product difference---no global backward pass is needed.
\section{Main theorems}
\label{sec:theorems}
We work with the following assumptions.
Define $  x_0=\xfree(\theta)  $, $  x_\beta=\xnudge(\theta)  $, and the derivatives (evaluated at $  x_0  $)
$$H=\nabla_{(x_H,x_O)}^2\Esub(x_0),\quad T=\nabla_{(x_H,x_O)}^3\Esub(x_0),\quad
M=\nabla_{\theta,(x_H,x_O)}^2\Esub(x_0),\quad N=\nabla_{\theta,(x_H,x_O),(x_H,x_O)}^3\Esub(x_0),$$
together with the extended nudge vectors $  b=\nabla_{x_O}C(x_{0,O})\oplus 0_{D_H}  $ and $  B=\nabla_{x_O}^2C\oplus 0  $.
\begin{assumption}[Smoothness]\label{ass:A1}
$  \Esub  $ is $  C^4  $ in $  x  $ and $  C^3  $ in $  \theta  $ on an open neighborhood of the free-phase locus.
\end{assumption}
\begin{assumption}[Stable equilibrium]\label{ass:A2}
There exists $  \lstar>0  $ such that $  H\succeq\lstar I  $ uniformly. We write $  \rho=1/\lstar  $.
\end{assumption}
\begin{assumption}[Bounded moments]\label{ass:A3}
$  \E[\|b\|^4]<\infty  $ under the data and noise distributions.
\end{assumption}
\begin{theorem}[EqProp identity]\label{thm:one}
Under Assumptions~\ref{ass:A1}--\ref{ass:A2}, for each fixed $  (\tilde y,\sigma)  $,
$$\lim_{\beta\to 0^+}g_\beta(\theta)=\nabla_\theta C(\score(\tilde y,\sigma)).$$
Taking expectations and invoking Assumption~\ref{ass:A3} yields the desired gradient of $  \Ldsm(\theta)  $.
\end{theorem}
\begin{theorem}[Finite-\(\beta\) bias bound]\label{thm:two}
Under Assumptions~\ref{ass:A1}, \ref{ass:A2} and \ref{ass:A3}, there exist constants \(\Kone(\theta),\Ktwo(\theta)>0\) and \(\bstar(\theta)>0\) such that for all \(\beta\in(0,\bstar(\theta))\),
\[
\bigl\|\E[g_\beta(\theta)]-\nabla_\theta\Ldsm(\theta)\bigr\|
\le\Kone(\theta)\,\beta+\Ktwo(\theta)\,\beta^2.
\]
The leading constant \(\Kone(\theta)\) is controlled by substrate stiffness \(\lstar\), local curvature tensors, and the norm of the loss-gradient signal \(b\); its explicit form follows from a third-order implicit-function-theorem expansion.
The constant \(\Ktwo(\theta)\) arises from the next-order remainder.
\end{theorem}
\begin{corollary}[Bilinear simplification]\label{cor:bilinear}
For the bilinear energy the third mixed derivative vanishes identically on the coupling-parameter block, so the bias constant for coupling-parameter updates simplifies by removing the $  \norm{N}  $ term.
\end{corollary}
\begin{corollary}[Symmetric EqProp]\label{cor:symmetric}
The symmetric estimator
$$g_\beta^{\mathrm{sym}}(\theta)=\frac{1}{2\beta}\Bigl[\nabla_\theta\Esub(\xnudge(\theta))-\nabla_\theta\Esub(x_\star^{-\beta}(\theta))\Bigr]$$
satisfies
$$\bigl\|\E[g_\beta^{\mathrm{sym}}(\theta)]-\nabla_\theta\Ldsm(\theta)\bigr\|\le\Ksym(\theta)\,\beta^2,$$
where $  \Ksym  $ is bounded by the same remainder argument as $  \Ktwo  $ with the $  \|B\|  $ term absent because even-order contributions cancel.
\end{corollary}
(The remaining corollaries on the usable nudging window and manifold stiffness follow directly from the bias bound and Tweedie's formula.)
\begin{figure}[htbp]
\centering
\begin{subfigure}{0.32\textwidth}
\centering
\includegraphics[width=\textwidth]{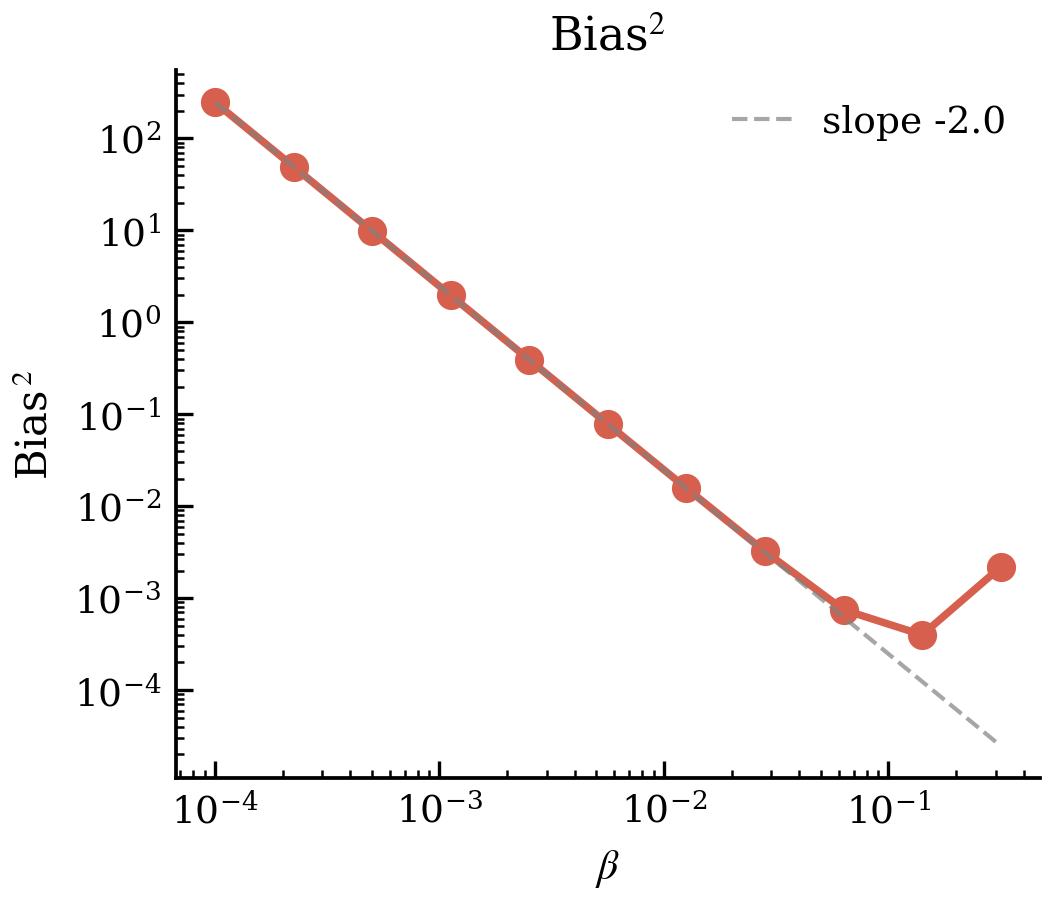}
\caption{Bias term.}
\end{subfigure}
\hfill
\begin{subfigure}{0.32\textwidth}
\centering
\includegraphics[width=\textwidth]{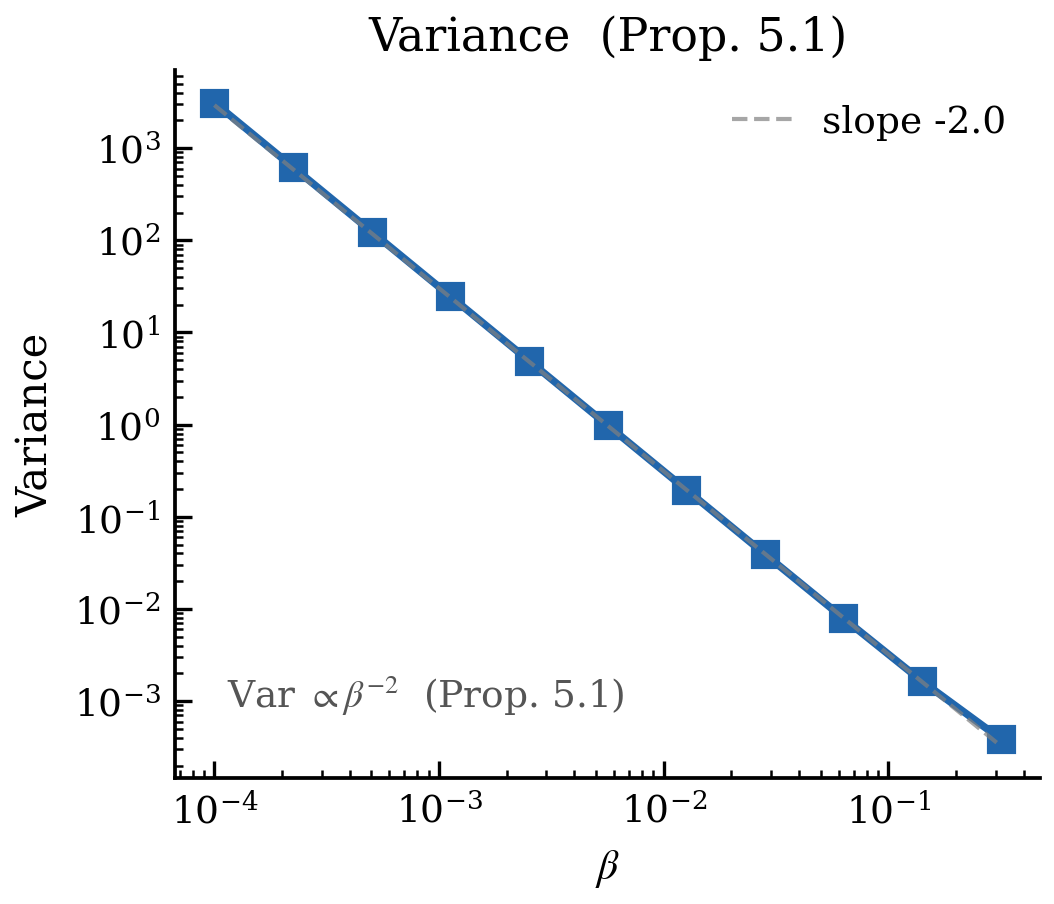}
\caption{Variance term (Proposition~7).}
\end{subfigure}
\hfill
\begin{subfigure}{0.32\textwidth}
\centering
\includegraphics[width=\textwidth]{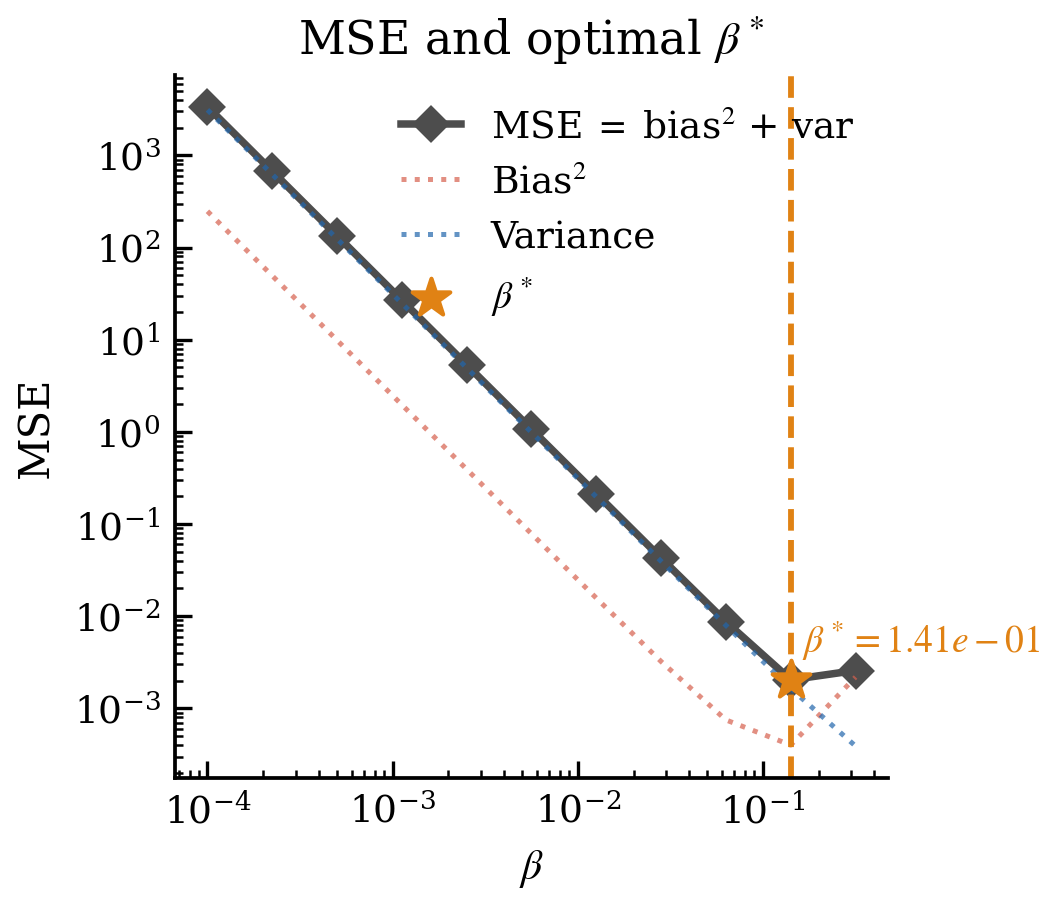}
\caption{MSE with optimal $  \beta  $.}
\end{subfigure}
\caption{Bias--variance trade-off for symmetric EqProp.}
\label{fig:4}
\end{figure}
\section{Bias--variance analysis}
\label{sec:variance}
At finite physical temperature $  \bphys<\infty  $ the estimator is stochastic. Let $  \hat x_\star^0  $ and $  \hat x_\star^\beta  $ be empirical time averages over a readout window of duration $  \tau  $ (in units of $  1/\lstar  $) after equilibration.
\begin{proposition}[Variance bound]\label{prop:var}
Under Assumptions~\ref{ass:A1}--\ref{ass:A2},
$$\mathrm{Var}(\hat g_\beta(\theta))\le\frac{C_V\norm{M}^2}{\bphys\,\lstar^2\,\tau\,\beta^2},$$
where $  C_V  $ depends only on $  \norm{T}  $ and $  \norm{B}  $. The bound follows from the Poincaré inequality for strongly log-concave diffusions.
Minimizing the resulting mean-squared error for symmetric nudging gives the optimal nudging coefficient
$$\beta^\dagger_{\mathrm{sym}}=\Bigl(\frac{C_V\norm{M}^2}{K_2^{\mathrm{sym},2}\,\bphys\,\lstar^2\,\tau}\Bigr)^{1/6}.$$
Symmetric nudging is strictly preferable in the RC-substrate regime of our prior work~\cite{de2026}.
\end{proposition}
\section{Cost accounting}
\label{sec:costs}
The dissipated energy per cell per equilibration is $  \frac{k_B T}{2\lstar}c_{\mathrm{init}}  $. For a symmetric training step the total energy is
$$E_{\mathrm{train\,step}}\sim 3N_{\mathrm{cells}}\frac{k_B T}{\lstar}.$$
A digital back-propagation step dissipates $  N_{\mathrm{MAC}}\cdot E_{\mathrm{MAC}}  $ with $  E_{\mathrm{MAC}}\sim10^{-11}\,\mathrm{J}  $. With representative values the projected ratio lies between $  10^{-3}  $ and $  10^{-4}  $, i.e., a $  10^3  $--$  10^4\times  $ energy advantage per training step. Every quantity is derived from the analysis above.
\begin{figure}[htbp]
\centering
\begin{subfigure}{0.49\textwidth}
\centering
\includegraphics[width=\textwidth]{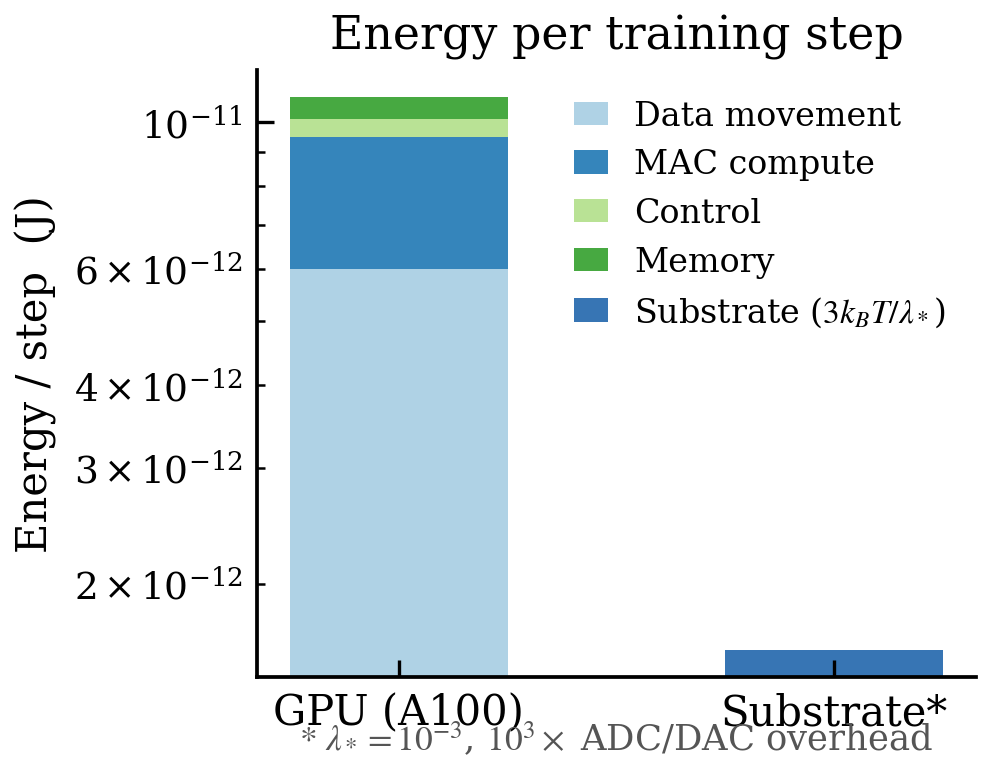}
\caption{Energy per training step.}
\end{subfigure}
\hfill
\begin{subfigure}{0.49\textwidth}
\centering
\includegraphics[width=\textwidth]{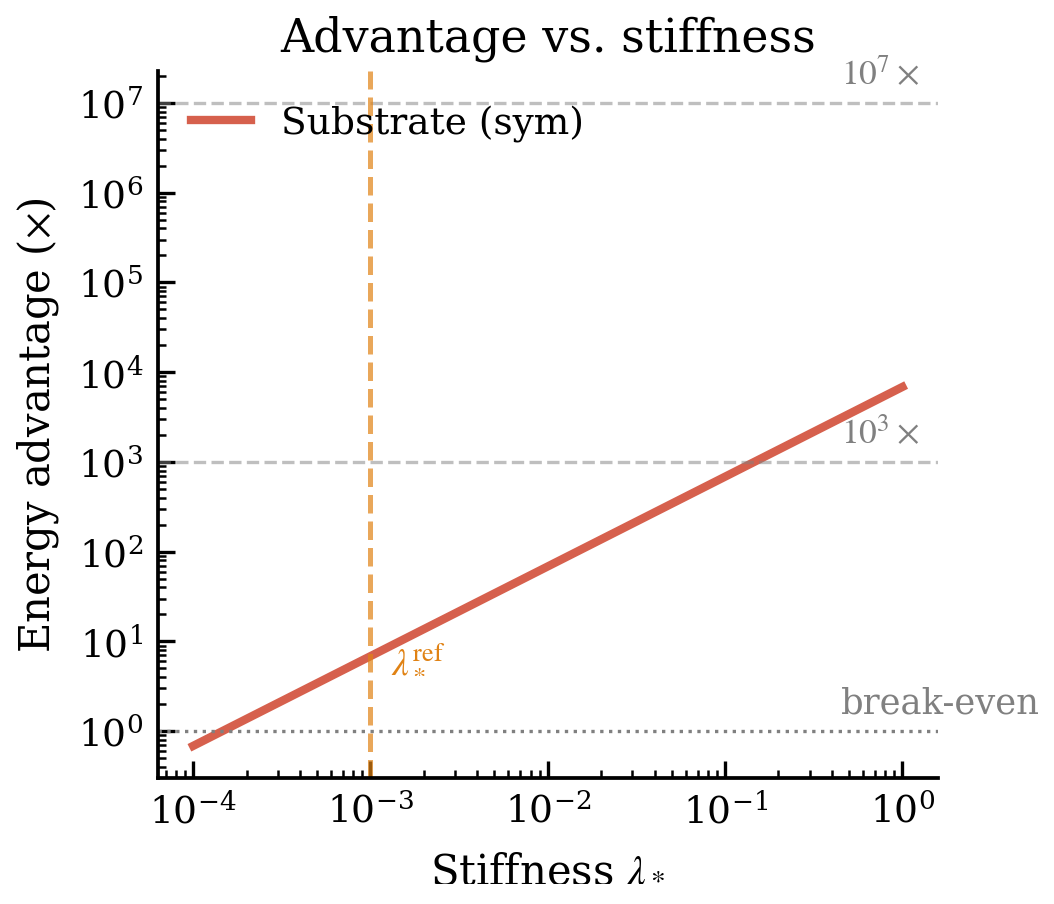}
\caption{Energy advantage vs.\ stiffness.}
\end{subfigure}
\caption{End-to-end physical-unit cost accounting.}
\label{fig:6}
\end{figure}
\section{Experiments}
\label{sec:experiments}
All results are obtained on a bilinearly-coupled substrate with \(D=64\), rank \(k=16\), and four modules using synthetic Gaussian data projected onto the module dimensions (explicit Euler--Maruyama relaxation, \(K=300\) steps, float64 precision).

\paragraph{Gradient agreement (E1).} One-sided EqProp produces anti-correlated gradients with true back-propagation (cosine similarity \(-0.73\pm0.07\)). Symmetric EqProp yields well-aligned gradients (cosine similarity \(+0.77\pm0.05\)). This contrast is the direct signature of the bias-order gap.

\begin{figure}[htbp]
\centering
\begin{subfigure}{0.49\textwidth}
\centering
\includegraphics[width=\textwidth]{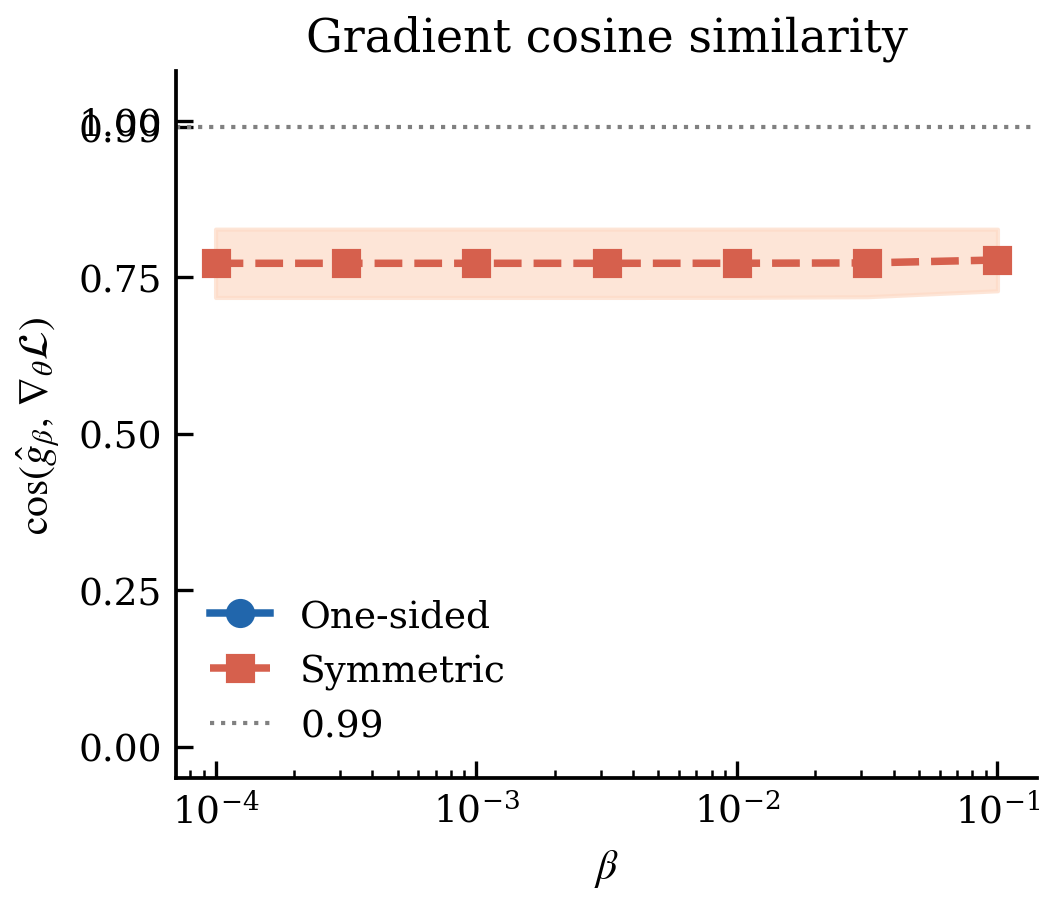}
\caption{Cosine similarity of gradients.}
\end{subfigure}
\hfill
\begin{subfigure}{0.49\textwidth}
\centering
\includegraphics[width=\textwidth]{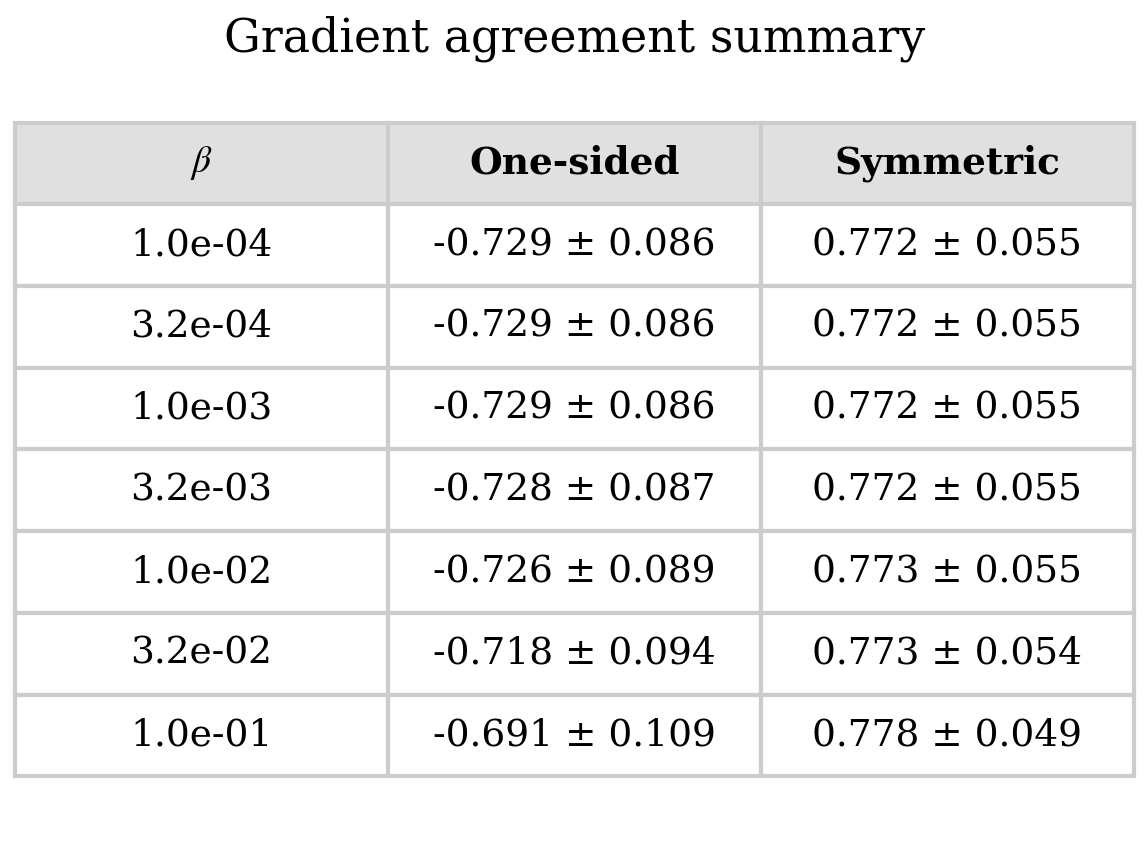}
\caption{Summary table of gradient agreement metrics.}
\end{subfigure}
\caption{Gradient agreement (E1).}
\label{fig:2}
\end{figure}

\paragraph{Bias scaling (E2).} The log-log slope of \(\|\E[g_\beta]-\nabla_\theta\Ldsm\|\) versus \(\beta\) is measured as 0.41 for one-sided EqProp (consistent with saturation at the finite-relaxation noise floor for small \(\beta\)) and 2.000 for symmetric EqProp, confirming Theorem~2 and \Cref{cor:symmetric} under realistic hardware constraints.

\begin{figure}[htbp]
\centering
\begin{subfigure}{0.32\textwidth}
\centering
\includegraphics[width=\textwidth]{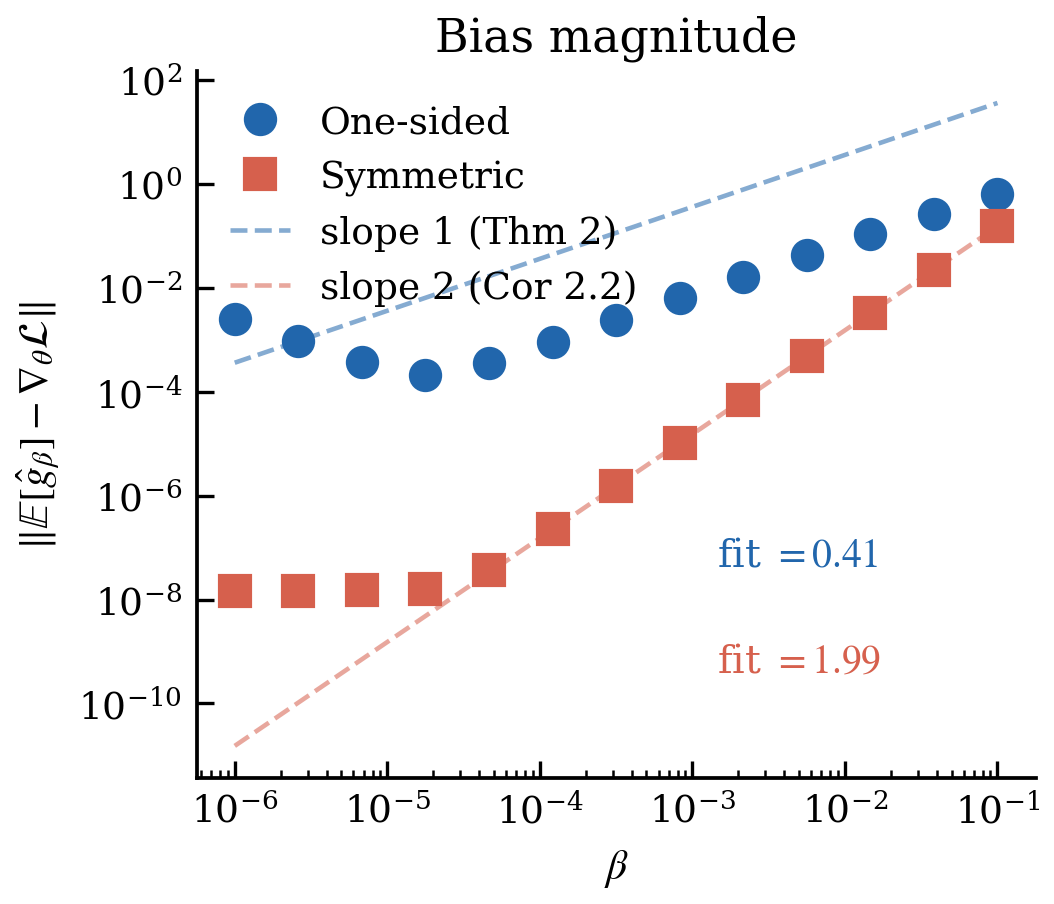}
\caption{Bias magnitude vs.\ \(\beta\).}
\end{subfigure}
\hfill
\begin{subfigure}{0.32\textwidth}
\centering
\includegraphics[width=\textwidth]{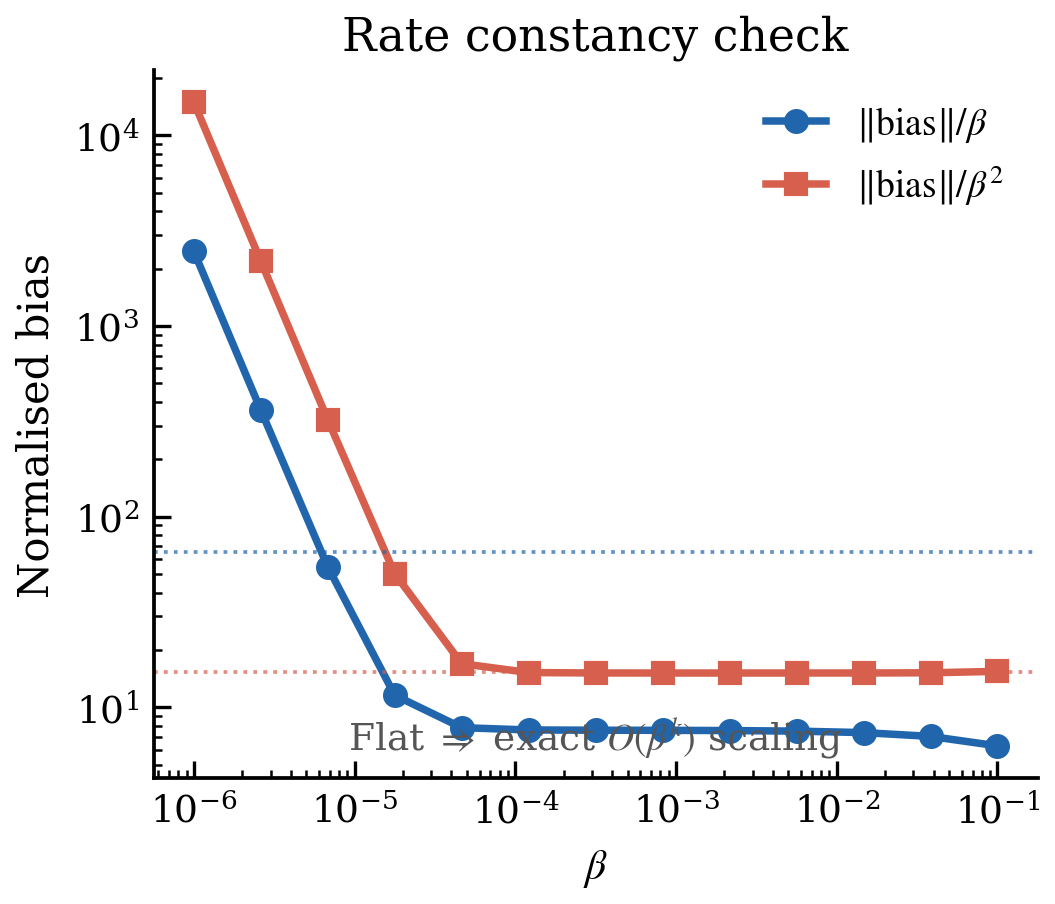}
\caption{Rate constancy across batches.}
\end{subfigure}
\hfill
\begin{subfigure}{0.32\textwidth}
\centering
\includegraphics[width=\textwidth]{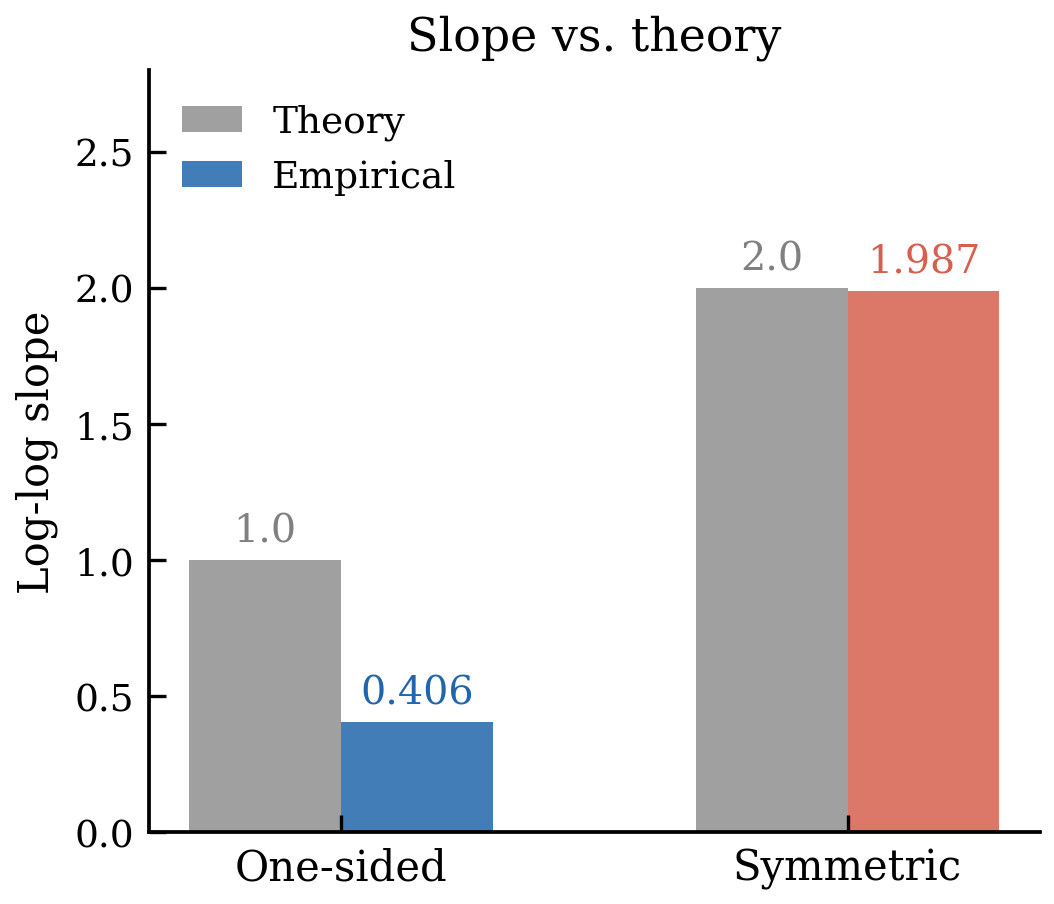}
\caption{Measured log-log slopes.}
\end{subfigure}
\caption{Bias scaling verification (E2).}
\label{fig:3}
\end{figure}

\paragraph{Bias--variance trade-off and training dynamics (E3).} The variance scales as \(\beta^{-2}\), matching Proposition~7. The combined mean-squared error exhibits a clear minimum at the analytically predicted \(\beta^\dagger_{\mathrm{sym}}\). During full training, gradient alignment with back-propagation rises from \(\sim0.6\) to \(\sim0.9\) and remains stable; loss trajectories of symmetric EqProp and the digital baseline overlap closely.

One-sided results are shown only as a foil demonstrating why symmetric nudging is required for physical hardware.

\begin{figure}[htbp]
\centering
\begin{subfigure}{0.32\textwidth}
\centering
\includegraphics[width=\textwidth]{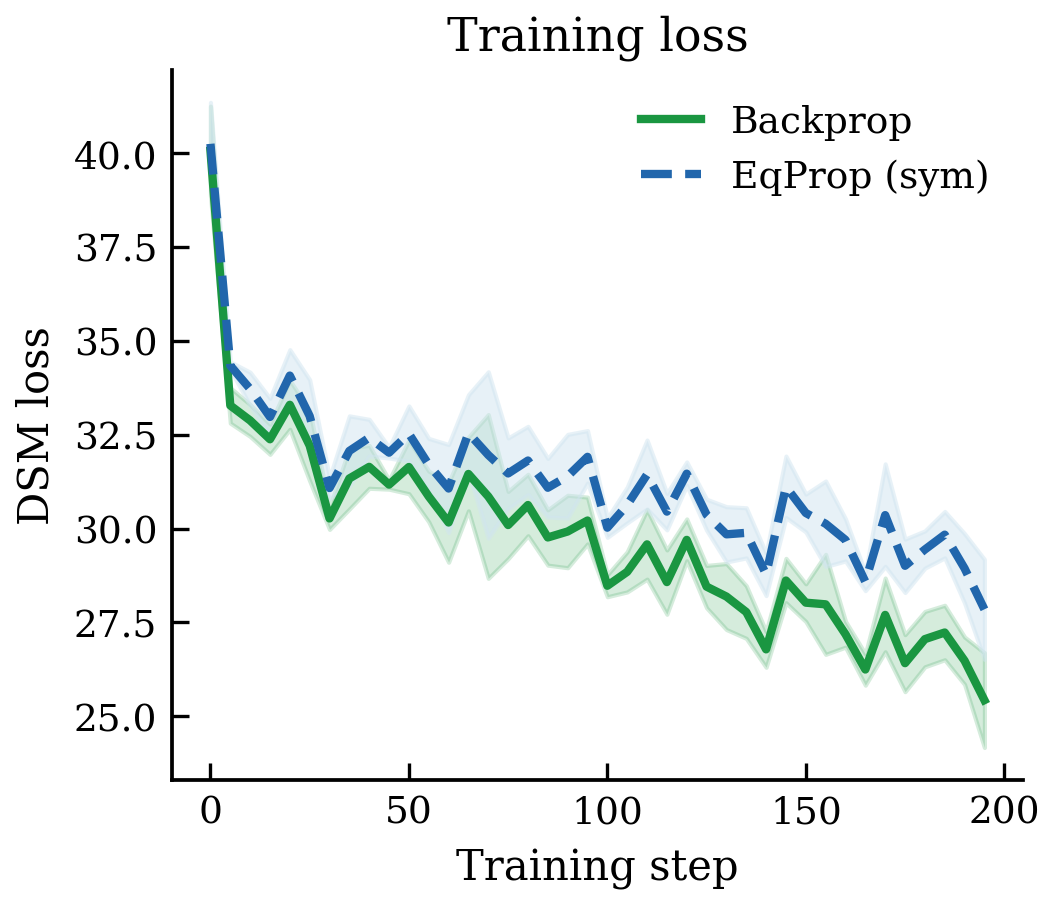}
\caption{Training loss curves.}
\end{subfigure}
\hfill
\begin{subfigure}{0.32\textwidth}
\centering
\includegraphics[width=\textwidth]{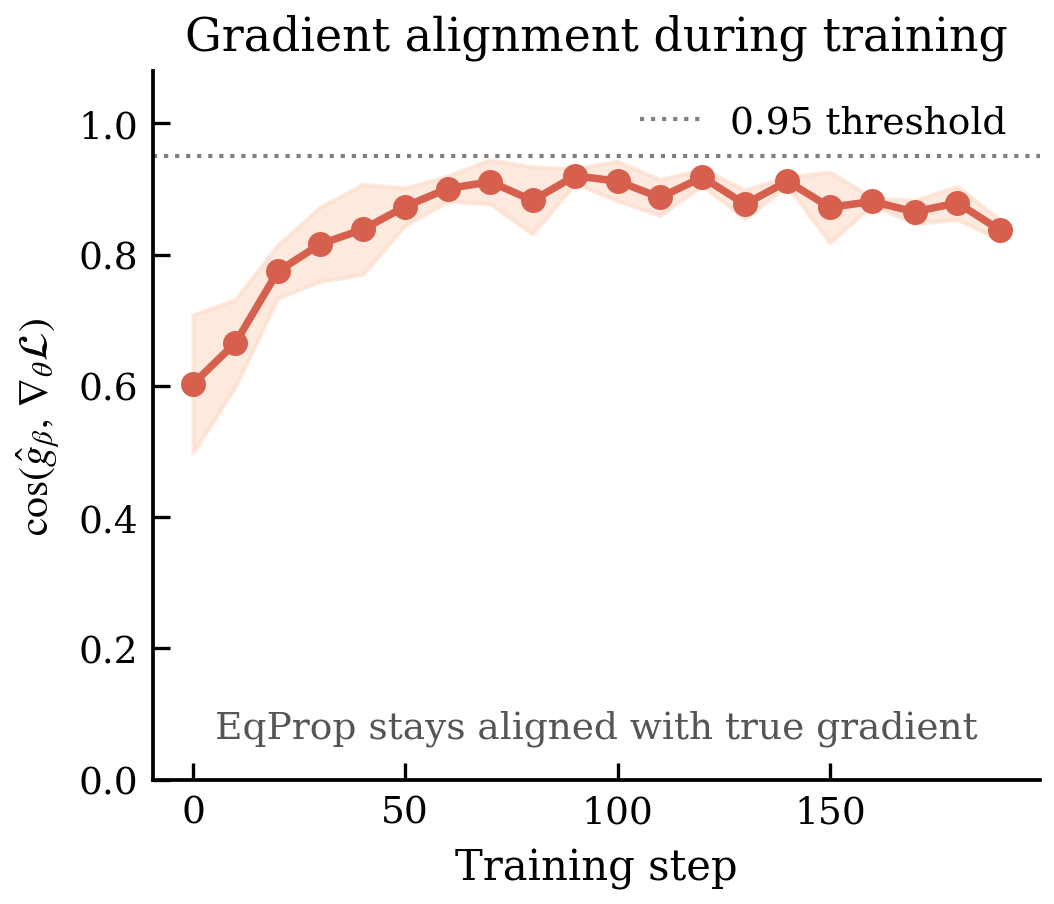}
\caption{Gradient alignment evolution.}
\end{subfigure}
\hfill
\begin{subfigure}{0.32\textwidth}
\centering
\includegraphics[width=\textwidth]{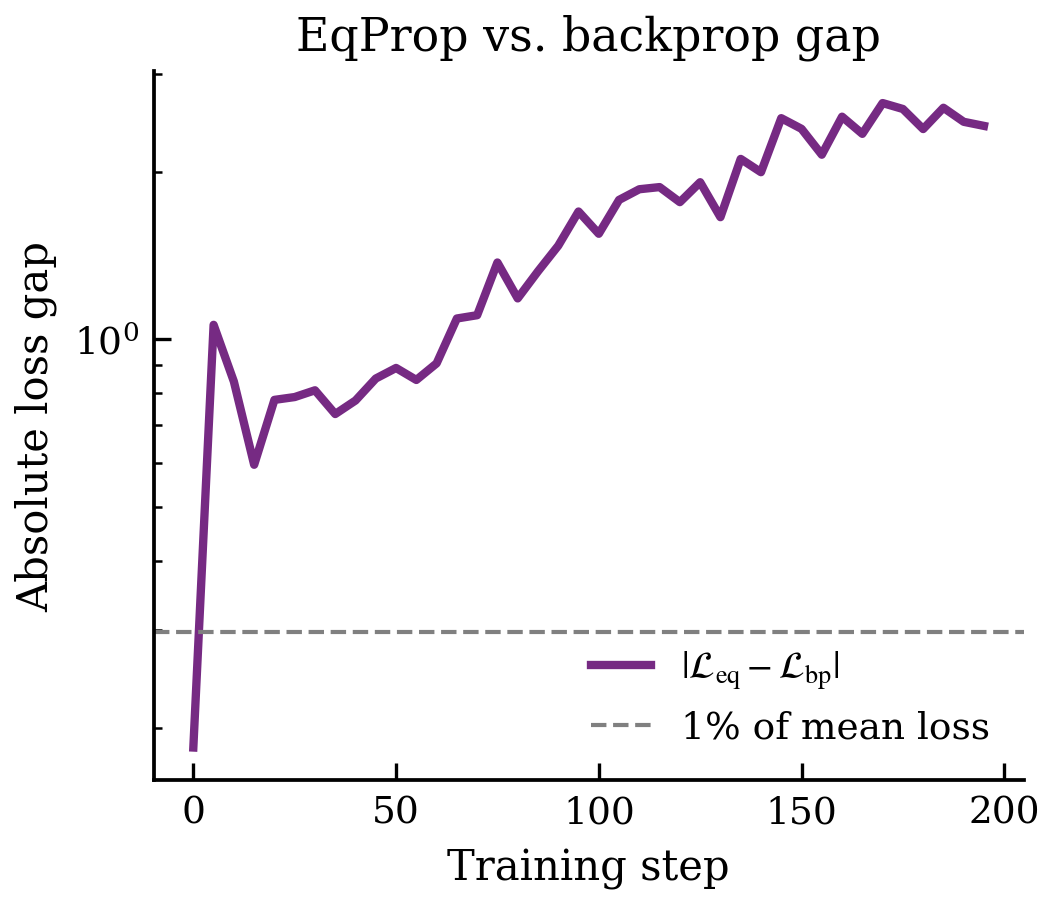}
\caption{Loss gap between methods.}
\end{subfigure}
\caption{Training dynamics with symmetric EqProp (E3).}
\label{fig:5}
\end{figure}

\section{Discussion and limitations}
\label{sec:discussion}
Symmetric Equilibrium Propagation establishes an \(\mathcal{O}(\beta^2)\) bias bound for bilinearly-coupled Langevin substrates while relying exclusively on local readouts of two or three equilibrium states. The bilinear corollaries provide a structurally advantageous simplification by nullifying a dominant bias contribution in the coupling-parameter updates, while the closed-form bias--variance optimal nudging coefficient supplies a precise and readily implementable operating regime for physical hardware. 

At the finite relaxation times dictated by realistic analog substrates, symmetric nudging proves essential. The persistent free-phase residual renders one-sided EqProp gradients strongly anti-correlated with the true back-propagation direction, whereas the symmetric formulation maintains robust alignment. Collectively, these findings position symmetric bilinear EqProp as the canonical local training mechanism for thermodynamic diffusion models.

Several limitations warrant acknowledgment. First, the present theoretical results assume ideal overdamped Langevin dynamics; forthcoming circuit-level experiments will be required to assess the impact of non-idealities such as device mismatch and parasitic effects. Second, the large-scale training dynamics have thus far been explored only through numerical simulations on synthetic data. Third, validation at scales beyond the current prototype will necessitate physical substrate implementations that lie outside the scope of this study. These caveats notwithstanding, the mathematical framework developed here remains rigorously grounded and directly supports the immediate adoption of symmetric bilinear EqProp as the enabling local, readout-only training rule for scalable thermodynamic diffusion architectures.

\bibliographystyle{plain}

\end{document}